\documentclass[letterpaper]{article} 
\usepackage{aaai24}  
\usepackage{times}  
\usepackage{helvet}  
\usepackage{courier}  
\usepackage[hyphens]{url}  
\usepackage{graphicx} 
\usepackage{natbib}  
\usepackage{caption} 
\usepackage{algorithm}
\usepackage{algorithmic}
\usepackage{xspace}
\usepackage{subfigure}
\usepackage{enumitem}
\usepackage{booktabs}
\usepackage{xcolor}
\usepackage{multirow}
\usepackage{color}
\usepackage{lipsum}
\usepackage{caption}
\usepackage{bm}
\usepackage{amsmath}
\usepackage{amsthm}
\usepackage{amsfonts}
\usepackage{amssymb}

\newtheorem{definition}{Definition}
\usepackage{newfloat}
\usepackage{listings}
\DeclareCaptionStyle{ruled}{labelfont=normalfont,labelsep=colon,strut=off} 
\floatstyle{ruled}
\newfloat{listing}{tb}{lst}{}
\floatname{listing}{Listing}
\title{GOODAT: Towards Test-time Graph Out-of-Distribution Detection}
\author{
    Luzhi Wang \textsuperscript{\rm 1}, Dongxiao He \textsuperscript{\rm 1}, 
    He Zhang \textsuperscript{\rm 2},
    Yixin Liu \textsuperscript{\rm 2},\\
    Wenjie Wang \textsuperscript{\rm 3},
    Shirui Pan \textsuperscript{\rm 4}\thanks{Corresponding author.},
    Di Jin \textsuperscript{\rm 1},
    Tat-Seng Chua \textsuperscript{\rm 3}
}
\affiliations{
    \textsuperscript{\rm 1}College of Intelligence and Computing, Tianjin University\\

    \textsuperscript{\rm 2}Faculty of Information Technology, Monash University\\
    \textsuperscript{\rm 3}School of Computing, National University of Singapore\\
    \textsuperscript{\rm 4}School of Information and Communication Technology, Griffith University\\
    \{wangluzhi, jindi, hedongxiao\}@tju.edu.cn, \{he.zhang1, yixin.liu\}@monash.edu, \\wenjiewang96@gmail.com, s.pan@griffith.edu.au, dcscts@nus.edu.sg
}

\usepackage{bibentry}

\begin{document}

\maketitle

\begin{abstract}
Graph neural networks (GNNs) have found widespread application in modeling graph data across diverse domains. While GNNs excel in scenarios where the testing data shares the distribution of their training counterparts (in distribution, ID), they often exhibit incorrect predictions when confronted with samples from an unfamiliar distribution (out-of-distribution, OOD). 
To identify and reject OOD samples with GNNs, recent studies have explored graph OOD detection, often focusing on training a specific model or modifying the data on top of a well-trained GNN. Despite their effectiveness, these methods come with heavy training resources and costs, as they need to optimize the GNN-based models on training data. Moreover, their reliance on modifying the original GNNs and accessing training data further restricts their universality. 
To this end, this paper introduces a method to detect \textbf{G}raph \textbf{O}ut-\textbf{o}f-\textbf{D}istribution \textbf{A}t \textbf{T}est-time (namely GOODAT), a \textit{data-centric}, \textit{unsupervised}, and \textit{plug-and-play} solution that operates independently of training data and modifications of GNN architecture. With a lightweight graph masker, GOODAT can learn informative subgraphs from test samples, enabling the capture of distinct graph patterns between OOD and ID samples. To optimize the graph masker, we meticulously design three unsupervised objective functions based on the graph information bottleneck principle, motivating the masker to capture compact yet informative subgraphs for OOD detection. 
Comprehensive evaluations confirm that our GOODAT method outperforms state-of-the-art benchmarks across a variety of real-world datasets.
\end{abstract}

\section{Introduction}
Graph neural networks (GNNs) are potent representation learning methods that focus on processing graph data \cite{wang2019exact, DBLP:journals/corr/abs-1903-05948}, and have been widely used in financial networks \cite{zheng2021heterogeneous}, binary code analysis \cite{wang2023contrastive}, sarcasm detection \cite{DBLP:conf/aaai/WangDJLW023, DBLP:conf/ijcai/Yu0WLW023}, etc. 
Generally, GNNs provide strong support for accurate prediction of downstream tasks by capturing the distribution of training data \cite{DBLP:conf/iclr/KipfW17, DBLP:conf/cikm/ZhangWY0WYP21}. However, when these well-trained GNN models are deployed in open-world scenarios, they inevitably encounter graph samples from unknown classes, the so-called graph ``Out-of-Distribution (OOD)'' data \cite{DBLP:conf/wsdm/LiuD0P23}. On the OOD data, well-trained GNNs may not be effective, as the features and distribution patterns of these OOD graphs are not exposed during GNN training \cite{pmlr-v202-bai23a}. 
This situation can lead to prediction errors when dealing with these unknown distribution samples, thereby reducing the reliability of GNNs. In this scenario, an ideal GNN model should possess the capability to identify and reject OOD samples, rather than misclassifying them as belonging to the in-distribution (ID) classes. 

\begin{figure}[t]
	\centering
	\includegraphics[width=0.4\textwidth]{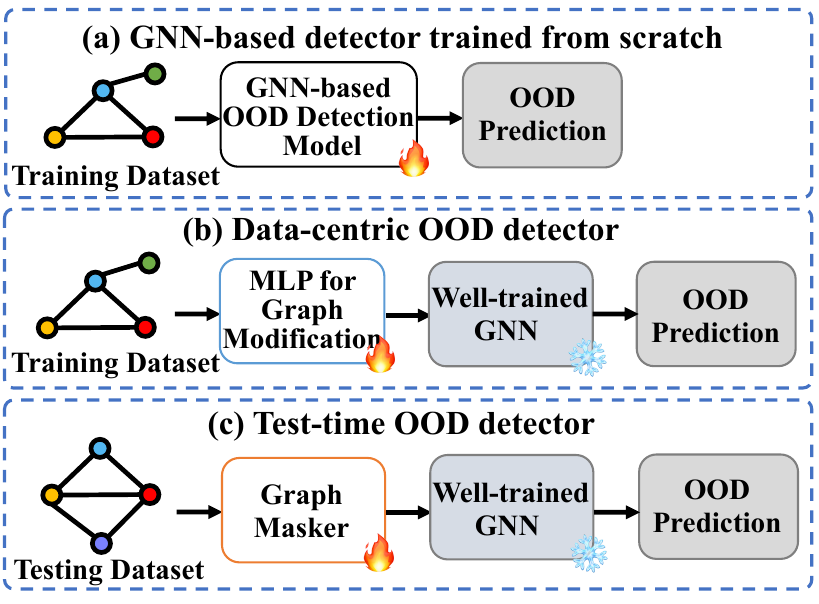} 
	\caption{Comparisons between GOODAT and other methods. 
 To detect OOD samples, \textbf{(a)} most GNN-based methods need to learn a detector from the training data \cite{DBLP:conf/wsdm/LiuD0P23}; \textbf{(b)} other data-centric methods learn an MLP to modify the training data while keeping the well-trained GNN fixed \cite{DBLP:conf/kdd/Guo0CLSD23}. 
    \textbf{(c)} In contrast, our test-time OOD detector directly works on the test data without needing to consult the training data and change the parameters of the well-trained GNN.
 }
	\label{intro}
\end{figure} 

To effectively identify OOD graph samples, various graph OOD detection approaches have emerged \cite{huang2022end, hoffmann2023open}. 
A subset of research~\cite{DBLP:conf/wsdm/LiuD0P23} centers around crafting GNN-based OOD detection models that are meticulously tailored for graph OOD detection tasks. While demonstrating effectiveness, they need to train additional GNNs from scratch, resulting in a heavy resource expenditure. 
Another type of methodology solves the graph OOD detection problem from a data-centric perspective \cite{zheng2023towards}, implementing downstream tasks by modifying data on top of well-trained GNNs. 
As a typical method, AAGOD~\cite{DBLP:conf/kdd/Guo0CLSD23} imposes a multi-layer perception (MLP) based parametric matrix on the adjacency matrix of each training graph. This is done without changing the parameters of the well-trained GNN, effectively widening the differences between OOD and ID graphs. By merely optimizing the MLP instead of retraining the GNN, AAGOD mitigates the effort required for model design and parameter training. 

Despite the effectiveness of the aforementioned methods, both types of existing graph OOD detection approaches rely on the training dataset, leading to several limitations. 
Firstly, training a task-specific GNN-based model for OOD detection usually requires significant \textit{computational resources and costs,} e.g., training an additional GNN from scratch \cite{DBLP:conf/iclr/00090DLTS23,DBLP:conf/nips/GuiLWJ22}. 
Secondly, in certain scenarios or platforms where model architectures are inaccessible due to privacy concerns \cite{zheng2023large}, making modifications or adjustments to the GNN architecture becomes impractical. Meanwhile, in some cases (e.g., federated learning \cite{tan2023heterogeneity}), the original training dataset may also be inaccessible \cite{tan2023federated}, further obstructing the training process of these OOD methods. 
In such instances, these aforementioned OOD detection methods cannot be applied, exposing their \textit{limited universality}. 

To address the above issues, we delve into the research problem of \textbf{test-time graph OOD detection}. 
Concretely, a test-time graph OOD detection method solely learns on testing data, without being dependent on training data. Moreover, the method is not expected to redesign the backbone model or add additional networks, ensuring its adaptability across diverse well-trained GNN models, irrespective of the characteristics of the training data. 
Nevertheless, developing such a test-time OOD detection method presents substantial complexities due to the following challenges. 
\textit{Challenge 1: Inconsistent learning objective.} Most well-trained GNNs are trained for specific graph learning tasks (e.g., graph classification) rather than OOD detection. In this case, how to align these pre-existing models with the target of OOD detection remains a difficulty. 
\textit{Challenge 2: Absence of labels.} During test time, the lack of graph labels poses a challenge in unsupervised detecting ID and OOD graphs, compelling the need to design an unsupervised model. 
\textit{Challenge 3: Unavailability of training data.} Constrained by the test-time setting, access to the training dataset is unfeasible. This scarcity of comprehensive knowledge about the original training data of the GNN model imposes significant limitations on our capacity to integrate an OOD detection model into GNNs. 

To solve the aforementioned challenges, we propose a novel data-centric method to detect \textbf{G}raph \textbf{O}ut-\textbf{O}f-\textbf{D}istribution \textbf{A}t \textbf{T}est-time, namely GOODAT. To address \textit{challenge 1}, we first construct a plug-and-play graph masker consisting of parameterized matrices. This enables us to compress informative subgraphs from the original input graphs, thereby indicating their ID or OOD nature. This lightweight masker can seamlessly integrate with any well-trained GNN, endowing it with the capability to detect OOD samples. 
To handle \textit{challenge 2}, we design three unsupervised loss functions based on the graph information bottleneck (GIB) principle, guiding the masker to capture compact yet sufficient subgraphs for distinguishing ID and OOD graphs. 
To deal with \textit{challenge 3}, we fully exploit the test data to capture the ID graph patterns of training data. Specifically, we operate under the assumption that all graphs in the test dataset are inherently ID. Guided by these surrogate ID labels, the OOD subgraph compressed by the ID label should significantly differ from the ID subgraph compressed by the ID label. This distinction serves as a reliable basis for effectively detecting OOD graphs. Fig. \ref{intro} shows the differences between GOODAT and other methods. 
To sum up, the main contributions of this paper are three-fold:
\begin{itemize}
    \item \textbf{New paradigm.} We pioneer the learning paradigm of test-time graph OOD detection, unveiling a fresh perspective. This innovative paradigm sheds light on \textit{lightweight}, \textit{training data-independent}, and \textit{plug-and-play} solutions for graph OOD detection, seamlessly applicable to any well-trained GNN models. 
    \item \textbf{Novel method.} We propose a simple yet effective method, GOODAT, to solve the test-time graph OOD detection problem. Leveraging the information bottleneck principle, GOODAT captures informative subgraphs from each input graph, thus enabling the accurate identification of OOD samples within the test dataset.
    \item \textbf{Extensive experiments.} We conduct experiments on multiple datasets and scenarios to verify the effectiveness and superiority of GOODAT. Experimental results show that GOODAT has achieved significant improvements in graph OOD detection tasks compared to baselines.
\end{itemize}

\begin{figure*}[tb]
	\centering
	\includegraphics[width=1\textwidth]{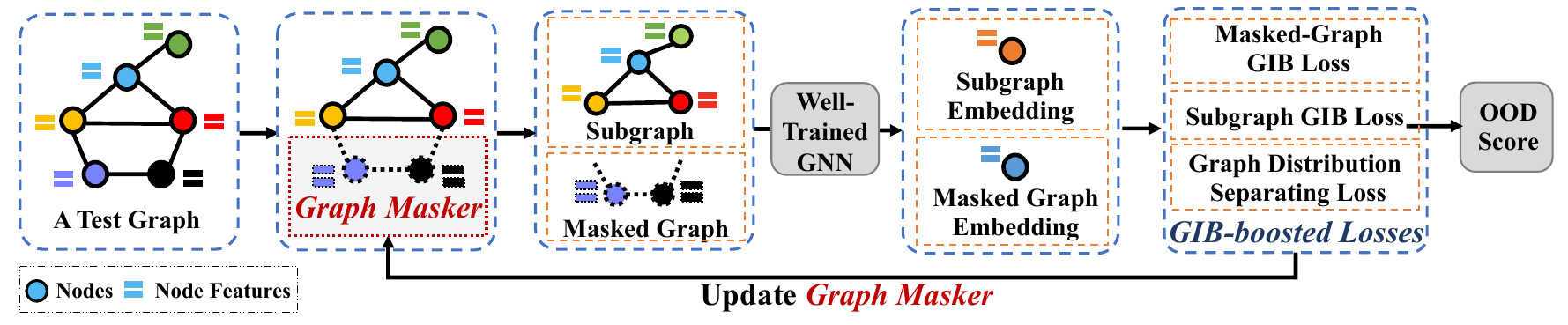} 
	\caption{Overview of GOODAT. In the GOODAT training process, a graph masker $M$ is applied on the input test graph $G$, consisting of two parameterized matrices.
This graph masker $M$ is trained by utilizing three GIB-boosted losses, taking the graph $G$ and its corresponding surrogate ID label $Y$ as inputs.
The informative subgraph $Z$ and the masked graph $Z'$ are obtained with the trainable parameters $M$ (e.g., $Z = G \odot M$).
During the inference phase of target GNNs, the OOD score of a test graph is obtained by the graph masker and GIB-boosted losses to infer if the input graph is an OOD graph.
}
	\label{flowchart}
 \vspace{-1mm}
\end{figure*} 
\section{Related Works}
\subsection{Graph Neural Networks}
With graph-related problems rising in various real-world scenarios \cite{ijcai2022p668,zheng2023gnnevaluator, WU2023106187}, GNNs have emerged as a powerful paradigm for tackling complex graph data \cite{liu2023learning, DBLP:conf/icml/ZhengZLZWP23, zheng2022rethinking}. 
GNNs have shown remarkable success across various domains \cite{DBLP:journals/corr/abs-2205-07424}, including social networks \cite{zheng2022rethinking}, anomaly detection \cite{liu2023towards}, binary code analysis \cite{DBLP:conf/ijcai/JinWZLJLP22}, and recommender systems \cite{DBLP:conf/www/JinWZSJLLP23, DBLP:journals/inffus/JinWZZDXP23}. 
Although many methods have been proposed to improve GNN performance \cite{DBLP:conf/iclr/KipfW17} \cite{DBLP:conf/iclr/XuHLJ19}, concerns have emerged about other aspects (e.g., robustness \cite{DBLP:journals/tkde/ZhangYZP23}, privacy \cite{DBLP:conf/icml/ZhangWWYXPY23,DBLP:conf/ndss/WuZYWXPY24}, fairness \cite{DBLP:journals/corr/abs-2301-12951}) of GNN models.
In the context of generalization, while GNNs perform well on ID data, they may perform poorly on OOD data \cite{DBLP:conf/wsdm/LiuD0P23}.

\subsection{Graph OOD Detection}
Graph OOD detection aims to detect whether the test graph is ID or OOD. Recently, many studies have been proposed \cite{zhao2020uncertainty, stadler2021graph, DBLP:conf/iclr/WuCYY23}. For example, GOOD-D \cite{DBLP:conf/wsdm/LiuD0P23} uses graph contrastive learning to provide an unsupervised view of graph OOD detection. 
GraphDE \cite{li2022graphde} models the graph generative process to learn latent environment variables for detection. 
OODGAT \cite{DBLP:conf/kdd/SongW22} utilizes a multi-head attention mechanism to compute node weights and transform them into edge weights, aiding in the identification of OOD nodes.
AAGOD \cite{DBLP:conf/kdd/Guo0CLSD23} introduces a data-centric framework that enlarges the differences between OOD and ID graphs.
However, the above methods rely on the training dataset to train OOD detection models, which is different from our GOODAT method.

\section{Preliminaries and Problem Definition}
\noindent\textbf{Graphs} and \textbf{Graph Maskers.} Given an undirected graph $G=(X, A)$, $X\in\mathbb{R}^{n \times d}$ represents the node feature matrix with $n$ nodes and each node has a feature dimension of $d$, and $A\in \mathbb{R}^{n\times n}$ indicates the adjacency matrix of $G$.
A label $Y$ is associated with $G$, where $Y =0$ indicates that the graph is ID, and $Y=1$ indicates that the graph is OOD. 
A graph masker is defined as ${M = (M_X, M_A)}$, where ${M_X \in \mathbb{R}^{n \times d}}$ and ${M_A \in \mathbb{R}^{n\times n}}$ are parameterized matrices for extracting the subgraph from the original graph $G$. We can modify the graph masker by gradient descent on $M_X$ and $M_A$. For example, we can obtain a subgraph ${Z = G \odot M = (X\odot M_X, A\odot M_A)}$ from $G$, where $\odot$ denotes the Hadamard product. 
Given $Z$, the remaining part of the test graph is donated as the masked graph ${Z' = (X-X\odot M_X, A-A\odot M_A)}$. For a given test graph $G=Z \cup Z'$, the overlap between $Z$ and $Z'$ is defined as $Z \cap Z'$, and $|Z \cap Z'|$ represents its size.

\noindent\textbf{Graph Information Bottleneck (GIB).} 
From the view of information theory, the information bottleneck aims at compressing the original information to obtain crucial information related to the label \cite{DBLP:conf/iclr/AlemiFD017}. 
As for graph information bottleneck, given a graph $G$ and its label $Y$, the graph information bottleneck aims to compress the information of $G$ to obtain a compressed graph $Z$, which maximizes the mutual information between $Y$ and $Z$ and minimizes the mutual information between $Z$ and $G$ \cite{DBLP:conf/iclr/YuXRBHH21}.
Specifically, assuming that $I(\cdot)$ indicates the Shannon mutual information, the GIB can be defined as \cite{wu2020graph}:
\begin{equation}
\begin{aligned}
\max_{Z}{ I(Y,Z) - \lambda I(G,Z) },
\end{aligned}  
\label{GIB}
\end{equation}
where $\lambda$ is a Lagrange multiplier.

\noindent\textbf{Test-time Graph OOD Detection.} 
For a test graph $G$, we assumed that it comes from the ID graph distribution or the OOD graph distribution.
We can define the test-time graph OOD detection task as:

\begin{definition}[Test-time graph OOD detection]
    Given a well-trained GNN $f$ and a graph $G$ from the test dataset, the test-time graph OOD detection aims to determine the source distribution of $G$ during the inference time of $f$ with an OOD detector $\texttt{D}$. Specifically, the objective of the detection task is:
\begin{equation}
\label{eq:ooddetection}
\texttt{Detection label}=
\begin{cases}
1\ (OOD), & if \ \texttt{D}(f, G) \geqslant \eta  \\ 
0\ (ID), & if \ \texttt{D}(f, G) < \eta \\
\end{cases}
\end{equation}
where $\eta$ is a threshold, and the parameters of $f$ are fixed during the OOD detection. 
\end{definition}

\begin{figure}[t]
	\centering
	\includegraphics[width=0.35\textwidth]{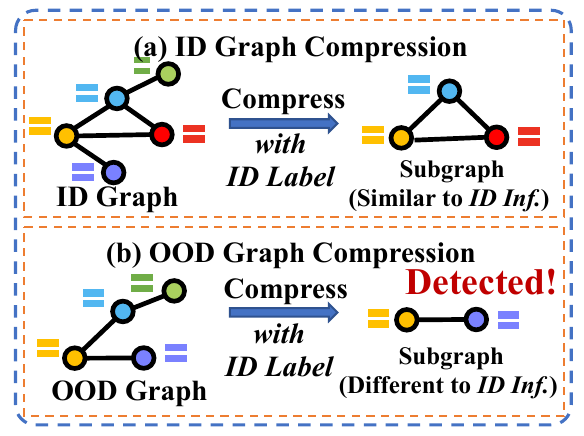} 
	\caption{The core idea of GOODAT. `Inf.' is the abbreviation of information.}
 \vspace{-3mm}
	\label{mainidea}
\end{figure} 
\vspace{-2mm}

\section{Methods}
Fig. \ref{flowchart} shows the overview of our GOODAT method, whose effectiveness in distinguishing ID and
OOD graphs is demonstrated below.
Given the surrogate label, GIB is employed to find the subgraph with the highest correlation to its label for each input graph. Hence, the subgraphs obtained from the ID and OOD graphs can be distinguishable, since they come from different distributions. 
During the test-time, a well-trained GNN $f$ tends to make right predictions on ID graphs while making wrong predictions on OOD graphs, since OOD graphs can be regarded as being in an unknown class. 
In this case, the subgraph of OOD graphs extracted by the GIB principle can be significantly different from the ones of ID graphs.
As shown in Fig. \ref{mainidea}, considering an ID graph $G^{I}$ with the right label prediction (e.g., $Y=0$), the subgraph obtained from GIB is denoted as $Z_{0}^{I}$. 
When the predicted label (e.g., $Y=0$) is wrong (e.g., for an OOD graph $G^{O}$), the extracted subgraph $Z_{0}^{O}$ obviously differentiates from $Z_{0}^{I}$ as $Z_{0}^{I}$ is not included in $G^{O}$. 
Otherwise, $G^{O}$ is not an OOD graph.
Such distinguishability between $Z_{0}^{I}$ and $Z_{0}^{O}$ can effectively separate ID and OOD graphs, enabling our method to execute the test-time OOD detection. 

In this paper, three different GIB-boosted loss functions are presented to enhance the capability of GIB on reasoning subgraphs that most correlate with labels.
The following are details on each of the three loss functions that play different roles in our optimization.

\subsection{Subgraph GIB Loss}
From the perspective of factual reasoning, the subgraph GIB loss facilitates the extraction of informative subgraph $Z$ that related to the predicted label. With the surrogate label $Y$ and obtaining the informative subgraph $Z = G \odot M$ by applying the graph masker $M=(M_X, M_A)$, we propose to utilize the following subgraph GIB loss to optimize the parameters $M_X$ and $M_A$ in $M$.
Specifically, we maximize the Shannon mutual information between the subgraph $Z$ (embedding from $f$) and the label $Y$, while minimizing the Shannon mutual information between the subgraph $Z$ and the original test graph $G$ (embedding from $f$).  
We seek the most informative and compressed subgraph representation by optimizing the following objective:
\begin{equation}\small
\max_{Z}I(Z,Y) - \alpha I(Z,G),
\label{mizy}
\end{equation}
where $\alpha$ is the Lagrange multiplier to balance the two components. 
Building on the work \cite{DBLP:conf/iclr/AlemiFD017}, we transfer $I(Z,Y)$ and $I(Z,G)$ into our loss function.

\noindent
\textbf{(1)} According to the definition of mutual information, we obtain $I(Z,Y)= \iint p(Z,Y)\log\frac{p(Y|Z)}{p(Y)} dYdZ$. For the item ${p( Y|Z) =\int \frac{p( Y|G) p( Z|G) p( G)}{p( Z)} dG}$, which depends on the Markov property \cite{DBLP:conf/aaai/Sun0P0FJY22} and is difficult to calculate, we propose using $q(Y|Z)$ to approximate $p(Y|Z)$. 
Specifically, we use the Kullback-Leibler divergence \cite{kullback1951information} to measure their distance and make them closer. 
Since ${D_{KL}( p( Y|Z) ||q( Y|Z)) \geqslant 0}$, we obtain
\begin{equation}
\small
    \int p( Y|Z)\log p( Y|Z) dy\geqslant \int p( Y|Z)\log q( Y|Z) dY.\\
    \label{6}
\end{equation}
According to Eq. (\ref{6}), the lower bound of $I( Z,Y)$ is:
\begin{equation}
\resizebox{0.47\textwidth}{!}{$
\begin{aligned}
I( Z,Y) &\geqslant \iint p( Y,Z)\log \frac{q( Y,Z)}{q(Z)} dZdY-\int p( Y)\log p( Y) dY. \\
 &\geqslant \iiint \frac{p( Y,G)p( Z,G)}{p( G)}\log \frac{q( Y,Z)}{q(Z)}dZdYdG.
\end{aligned}$}
\label{middle}
\end{equation}
\noindent
\textbf{(2)} For $I(Z,G) = \iint p( Z,G)\log\frac{p( Z,G)}{p( Z)p(G)} dZdG$, where it is not easy to directly calculate $p(Z)$, we propose to use a learnable $\phi(Z)$ to approximate $p(Z)$. Similarly, since $D_{KL}( p( Z)| |\phi ( Z))\geqslant 0$, we can get an upper bound of $I(Z,G)$:
\begin{equation}
\small
\begin{aligned}
    I( Z,G) \leqslant \iint p( Z,G)\log\frac{p( Z,G)}{\phi ( Z)p(G)} dGdZ.
\end{aligned}
\end{equation}
By combining the above inequalities about $I(Z,Y)$ and $I(Z,G)$, we derive a lower bound of $I(Z,Y) - \alpha I(Z,G)$. Specifically, 
\begin{equation}
\small
\begin{aligned}
        &I( Z,Y) -\alpha I( Z,G) \\
        &\geqslant \iiint \frac{p( Y,G)p( Z,G)}{p( G)}\log \frac{q( Y,Z)}{q(Z)}dZdYdG \\
        &-\alpha \iint p( Z,G)\log\frac{p( Z,G)}{\phi ( Z)p(G)} dGdZ.
\end{aligned}
\end{equation}
where $\alpha$ is a hyperparameter that balances informativeness and compression.
Thus, instead of directly solving $\max_{Z}I( Z,Y) -\alpha I( Z,G)$, we employ the following subgraph GIB loss $\mathcal{L}_s$:
\begin{equation}\small
\begin{aligned}
        \mathcal{L}_{s}  = &\frac{1}{N}\sum _{i=1}^{N}(\mathbb{E}_{\xi \sim p( \xi )}( -\log q( Y_{i} |Z_{i} )) +\alpha D_{KL}[ p( Z_{i} |G_{i}) ||\phi ( Z_{i})]), \\
        &\approx \mathcal{L}_{cls}(q( Y_{i} | Z_{i}), Y_i)+ \alpha D_{KL}[ p( Z_{i} |G_{i}) ||\phi ( Z_{i})]), 
\end{aligned}
\label{ls}
\end{equation}
where $N$ represents the number of graphs, $\mathcal{L}_{cls}$ indicates the classification loss. We use the Gaussian distribution to approximate $q$ and $\phi$. 
Fig. \ref{threeloss} (a) illustrates the impact of subgraph GIB loss on the subgraph compression process.

\subsection{Masked Graph GIB Loss}

From the perspective of counterfactual reasoning, the masked graph GIB loss is used to obtain the masked graph $Z'$, which ensures the irrelevance between the $Z'$ and the label.
Considering that $Z'$ is regarded as the noise part of the graph $G$, the optimization objective from the GIB perspective can be formulated as:
\begin{equation}\small
    \max_{Z'}\beta I( Z',G) -I( Z',Y),
\end{equation}
where $\beta$ is a Lagrange multiplier.
Similar to the derivation of the subgraph GIB loss, the masked graph GIB loss $\mathcal{L}_{m}$ can be defined as:
\begin{equation}\small
    \begin{aligned}
        \mathcal{L}_{m} 
        &=\frac{1}{N}\sum _{i=1}^{N}(\mathbb{E}_{\xi \sim p( \xi )}(\log q( Y_{i} |Z_{i} ') -\beta D_{KL}[ p( Z_{i} '|G_{i}) ||\phi ( Z_{i} ')]), \\
        &\approx -\mathcal{L}_{cls}(q( Y_{i} |Z_{i} '), Y_i)-\beta D_{KL}[ p( Z_{i} '|G_{i}) ||\phi ( Z_{i} ')]).
    \end{aligned}
    \label{lm}
\end{equation}
We use the Gaussian distribution to approximate $q$ and $\phi$. Fig. \ref{threeloss} (b) illustrates the masked graph GIB loss.

According to an existing study \cite{DBLP:conf/www/TanGFGX0Z22}, only considering the subgraph GIB loss $\mathcal{L}_{s}$ leads to sufficient but not necessary $Z$, while exclusively using the masked graph GIB loss $\mathcal{L}_{m}$ tends to generate necessary but not sufficient $Z$. To overcome this limitation (i.e., a large $|Z \cap Z'|$), employing $\mathcal{L}_{s}$ and $\mathcal{L}_{m}$ simultaneously helps to dig out sufficient and necessary $Z$ (i.e., limited $|Z \cap Z'|$).

\subsection{Graph Distribution Separating Loss}
To further reduce the overlapping size (i.e., $|Z \cap Z'|$), we propose the graph distribution separating loss to separate the distributions of $Z$ and $Z'$ from the statistical view,
where $Z$ and $Z'$ are supposed to own a small joint probability density. 
Sklar \cite{sklar1959fonctions} declares that the joint distribution of $N$ random variables can be decomposed into the respective marginal distributions of the $N$ variables and a Copula function \cite{sklar1973random}, so as to separate the randomness and coupling of the variables. 
In this paper, we assume that $Z$ and $Z'$ have the following Gaussian form of marginal distributions, i.e.,
\begin{equation}\small
\begin{aligned}
p_{Z}( Z) = \frac{1}{\sqrt{2\pi } |\sigma_1 |} e^{\frac{-Z^{2}}{2\sigma_1 ^{2}}} ,
p_{Z'}( Z') = \frac{1}{\sqrt{2\pi } |\sigma_2|} e^{\frac{-Z'^{2}}{2\sigma_2^{2}}},
\end{aligned}
\end{equation}
where $\sigma_1$ and $\sigma_2$ represent the standard deviation of $Z$ and $Z'$, respectively.

\begin{figure}[t]
	\centering
	\includegraphics[width=0.35\textwidth]{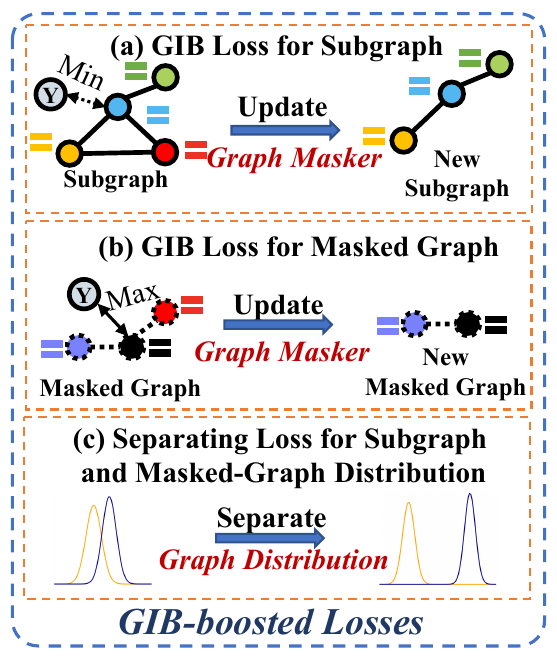} 
	\caption{Illustrations of the three GIB-boosted losses in GOODAT.}
	\label{threeloss}
\end{figure}

According to the Copula theory \cite{durante2013topological}, there must be a function that couples the marginal probabilities. Therefore, the joint probability function of $Z$ and $Z'$ can be defined as: 
\begin{equation}\small
\begin{aligned}
    \mathcal{L}_{d} = \frac{1}{2\pi |\sigma_1 \sigma_2|\sqrt{1-\rho ^{2}}} e^{\left[ -\frac{1}{2\left( 1-\rho ^{2}\right)} \times \left[\frac{Z^{2}}{\sigma_1 ^{2}} -2\rho \frac{Z\times Z'}{|\sigma_1 \sigma_2|} +\frac{Z'^{2}}{\sigma_2^{2}}\right]\right]},
\end{aligned}
\end{equation}
where $\rho$ indicates the correlation coefficient. 
Fig. \ref{threeloss} (c) illustrates how the graph distribution separating loss influences the distinguishability of graph distributions. 

\subsection{Training of GOODAT and OOD Graph Detection}
By integrating the above three loss functions together, the overall loss $\mathcal{L}_{g}$ for training $M$ can be expressed as:
\begin{equation}
    \mathcal{L}_{g} = \mathcal{L}_{s}+\mathcal{L}_{m}+\mathcal{L}_{d}.
\end{equation}
Note that in our method, all parameters in the target GNN $f$ are fixed, and we employ the surrogate label $Y$ and embeddings of $G$, $Z$, and $Z'$ to train the mask generator $M$. 

When implementing OOD detection, we employ the subgraph GIB loss (Eq. \ref{ls}) as the graph OOD detection score. Specifically, for a test graph $G$, we first use the well-trained graph masker $M$ to obtain its informative subgraph $Z=G\odot M$.
Next, we obtain the GIB loss on $Z$ w.r.t. $\mathcal{L}_{s}$, which is used to determine whether $G$ is an OOD graph by Eq. (\ref{eq:ooddetection}).

\section{Experiments}
\begin{table*}[t]
\centering
\caption{OOD detection results in terms of AUC score ($\%$). The best results are highlighted with \textbf{bold}.} 
\label{oodd}
\resizebox{1\textwidth}{!}{
\begin{tabular}{lccccccccccc}
\toprule
ID dataset & BZR & PTC-MR & AIDS & ENZYMES & IMDB-M & Tox21 & FreeSolv & BBBP & ClinTox & Esol & \textit{Avg.}  \\
\cmidrule{1-11}
OOD dataset & COX2 & MUTAG & DHFR & PROTEIN & IMDB-B & SIDER & ToxCast & BACE & LIPO & MUV  &\textit{Rank}\\
\midrule
PK-LOF      & $42.22{\scriptstyle\pm8.39}$ & $51.04{\scriptstyle\pm6.04}$ & $50.15{\scriptstyle\pm3.29}$ & $50.47{\scriptstyle\pm2.87}$ & $48.03{\scriptstyle\pm2.53}$ & $51.33{\scriptstyle\pm1.81}$ & $49.16{\scriptstyle\pm3.70}$ & $53.10{\scriptstyle\pm2.07}$ & $50.00{\scriptstyle\pm2.17}$ & $50.82{\scriptstyle\pm1.48}$ & $9.9$ \\
PK-OCSVM    & $42.55{\scriptstyle\pm8.26}$ & $49.71{\scriptstyle\pm6.58}$ & $50.17{\scriptstyle\pm3.30}$ & $50.46{\scriptstyle\pm2.78}$ & $48.07{\scriptstyle\pm2.41}$ & $51.33{\scriptstyle\pm1.81}$ & $48.82{\scriptstyle\pm3.29}$ & $53.05{\scriptstyle\pm2.10}$ & $50.06{\scriptstyle\pm2.19}$ & $51.00{\scriptstyle\pm1.33}$ & $10.0$ \\
PK-iF & $51.46{\scriptstyle\pm1.62}$ & $54.29{\scriptstyle\pm4.33}$ & $51.10{\scriptstyle\pm1.43}$ & $51.67{\scriptstyle\pm2.69}$ & $50.67{\scriptstyle\pm2.47}$ & $49.87{\scriptstyle\pm0.82}$ & $52.28{\scriptstyle\pm1.87}$ & $51.47{\scriptstyle\pm1.33}$ & $50.81{\scriptstyle\pm1.10}$ & $50.85{\scriptstyle\pm3.51}$ & $8.1$ \\
WL-LOF   &   $48.99{\scriptstyle\pm6.20}$ & $53.31{\scriptstyle\pm8.98}$ & $50.77{\scriptstyle\pm2.87}$ & $52.66{\scriptstyle\pm2.47}$ & $52.28{\scriptstyle\pm4.50}$ & $51.92{\scriptstyle\pm1.58}$ & $51.47{\scriptstyle\pm4.23}$ & $52.80{\scriptstyle\pm1.91}$ & $51.29{\scriptstyle\pm3.40}$ & $51.26{\scriptstyle\pm1.31}$ & $7.2$ \\
WL-OCSVM    & $49.16{\scriptstyle\pm4.51}$ & $53.31{\scriptstyle\pm7.57}$ & $50.98{\scriptstyle\pm2.71}$ & $51.77{\scriptstyle\pm2.21}$ & $51.38{\scriptstyle\pm2.39}$ & $51.08{\scriptstyle\pm1.46}$ & $50.38{\scriptstyle\pm3.81}$ & $52.85{\scriptstyle\pm2.00}$ & $50.77{\scriptstyle\pm3.69}$ & $50.97{\scriptstyle\pm1.65}$ & $8.1$ \\
WL-iF  & $50.24{\scriptstyle\pm2.49}$ & $51.43{\scriptstyle\pm2.02}$ & $50.10{\scriptstyle\pm0.44}$ & $51.17{\scriptstyle\pm2.01}$ & $51.07{\scriptstyle\pm2.25}$ & $50.25{\scriptstyle\pm0.96}$ & $52.60{\scriptstyle\pm2.38}$ & $50.78{\scriptstyle\pm0.75}$ & $50.41{\scriptstyle\pm2.17}$ & $50.61{\scriptstyle\pm1.96}$ & $9.3$ \\
\midrule
InfoGraph-iF & $63.17{\scriptstyle\pm9.74}$ & $51.43{\scriptstyle\pm5.19}$ & $93.10{\scriptstyle\pm1.35}$ & $60.00{\scriptstyle\pm1.83}$ & $58.73{\scriptstyle\pm1.96}$ & $56.28{\scriptstyle\pm0.81}$ & $56.92{\scriptstyle\pm1.69}$ & $53.68{\scriptstyle\pm2.90}$ & $48.51{\scriptstyle\pm1.87}$ & $54.16{\scriptstyle\pm5.14}$ & $5.4$ \\
InfoGraph-MD & $\mathbf{86.14{\scriptstyle\pm6.77}}$ & $50.79{\scriptstyle\pm8.49}$ & $69.02{\scriptstyle\pm11.67}$ & $55.25{\scriptstyle\pm3.51}$ & $\mathbf{81.38{\scriptstyle\pm1.14}}$ & $59.97{\scriptstyle\pm2.06}$ & $58.05{\scriptstyle\pm5.46}$ & $70.49{\scriptstyle\pm4.63}$ & $48.12{\scriptstyle\pm5.72}$ & $77.57{\scriptstyle\pm1.69}$ & $4.5$ \\
GraphCL-iF & $60.00{\scriptstyle\pm3.81}$ & $50.86{\scriptstyle\pm4.30}$ & $92.90{\scriptstyle\pm1.21}$ & $61.33{\scriptstyle\pm2.27}$ & $59.67{\scriptstyle\pm1.65}$ & $56.81{\scriptstyle\pm0.97}$ & $55.55{\scriptstyle\pm2.71}$ & $59.41{\scriptstyle\pm3.58}$ & $47.84{\scriptstyle\pm0.92}$ & $62.12{\scriptstyle\pm4.01}$ & $5.7$ \\
GraphCL-MD & $83.64{\scriptstyle\pm6.00}$ & $73.03{\scriptstyle\pm2.38}$ & $93.75{\scriptstyle\pm2.13}$ & $52.87{\scriptstyle\pm6.11}$ & $79.09{\scriptstyle\pm2.73}$ & $58.30{\scriptstyle\pm1.52}$ & $60.31{\scriptstyle\pm5.24}$ & $75.72{\scriptstyle\pm1.54}$ & $51.58{\scriptstyle\pm3.64}$ & $78.73{\scriptstyle\pm1.40}$ & $2.6$ \\
\midrule
GTrans & $55.17{\scriptstyle\pm5.04}$ & $62.38{\scriptstyle\pm2.36}$ & $60.12{\scriptstyle\pm1.98}$ & $49.94{\scriptstyle\pm5.67}$ & $51.55{\scriptstyle\pm2.90}$ & $61.67{\scriptstyle\pm0.73}$ & $50.81{\scriptstyle\pm3.03}$ & $64.02{\scriptstyle\pm2.10}$ & $58.54{\scriptstyle\pm2.38}$ & $76.31{\scriptstyle\pm3.85}$ & $5.5$ \\
\midrule
AAGOD-GIN\ensuremath{_S+} & $76.75$ & $-$ & $-$ & $66.22$ & $59.00$ & $64.26$  & $-$ & $67.80$  & $-$ & $-$ & $-$ \\
AAGOD-GIN\ensuremath{_L+} & $76.00$ & $-$ & $-$ & $65.89$ & $62.70$ & $57.59$  & $-$ & $57.13$  & $-$ & $-$ & $-$ \\
\midrule
Ours & $82.16{\scriptstyle\pm0.15}$ & $\mathbf{81.84{\scriptstyle\pm0.57}}$ & $\mathbf{96.43{\scriptstyle\pm0.25}}$ & $\mathbf{66.29{\scriptstyle\pm1.54}}$ & $79.03{\scriptstyle\pm0.03}$ & $\mathbf{68.92{\scriptstyle\pm0.01}}$ & $\mathbf{68.83{\scriptstyle\pm0.02}}$ & $\mathbf{77.07{\scriptstyle\pm0.03}}$ & $\mathbf{62.46{\scriptstyle\pm0.54}}$ & $\mathbf{85.91{\scriptstyle\pm0.27}}$ & $\mathbf{1.4}$ \\
\bottomrule
\end{tabular}}
\end{table*}

\subsection{Experimental Setup}
\subsubsection{Datasets}
In this paper, we utilize the evaluation protocol proposed by~\cite{DBLP:conf/wsdm/LiuD0P23} which encompasses a graph OOD detection benchmark and a graph anomaly detection benchmark. The graph OOD detection benchmark contains $8$ pairs of molecule datasets, $1$ pair of bioinformatics datasets, and $1$ pair of social network datasets. Each dataset pair within the same field exhibits a moderate domain shift. Following the setting in previous studies~\cite{DBLP:conf/wsdm/LiuD0P23,DBLP:conf/kdd/Guo0CLSD23}, we allocate $90\%$ of ID samples for training GNNs, while the remaining $10\%$ of ID samples and an equal number of OOD samples constitute the test set. The graph anomaly detection benchmark comprises $15$ datasets from TU benchmark \cite{DBLP:journals/corr/abs-2007-08663}. Anomalies encompass samples of the minority or real anomalous class, with the rest categorized as normal data.
In our test-time OOD detection setting, we use the test set of each dataset as the input of GOODAT for test-time training.

\subsubsection{Baselines $\&$ Settings}
We compared GOODAT with $13$ mainstream baseline methods for graph OOD detection, which 
can be divided into four categories. 
\begin{itemize}
    \item \textbf{Graph Kernels + Detectors.} This type of method uses graph kernels as the pre-train function to learn graph embedding, and then inputs these graph embeddings to OOD/anomaly detector. Graph kernels include the Weisfeiler-Lehman kernel (WL) \cite{DBLP:journals/jmlr/ShervashidzeSLMB11} and propagation kernel (PK) \cite{DBLP:journals/ml/NeumannGBK16}. OOD/anomaly detectors include local outlier factor (LOF) \cite{DBLP:conf/sigmod/BreunigKNS00}, one-class SVM (OCSVM) \cite{DBLP:journals/jmlr/ManevitzY01}, and isolation forest (iF) \cite{DBLP:conf/icdm/LiuTZ08}.
    \item \textbf{Graph Neural Networks + Detectors.} This type of method employs graph-level self-supervised GNN models as pre-trained models alongside OOD detectors to accomplish the detection. The self-supervised GNN models include InfoGraph \cite{DBLP:conf/iclr/SunHV020} and GraphCL \cite{DBLP:conf/nips/YouCSCWS20}. The detectors include iF and Mahalanobis distance (MD) \cite{DBLP:conf/iclr/SehwagCM21}.
    \item \textbf{Test-time Training Methods.} This type of method achieves graph OOD generalization at test-time, without any modification to the parameters of well-trained GNNs. The most recent method that falls into this category is GTrans \cite{DBLP:conf/iclr/00090DLTS23}. Since GTrans is not explicitly designed for graph OOD detection, its loss value is adopted as the OOD score in our experiments.
    \item \textbf{Data-centric Methods.} One quintessential method of this category is AAGOD \cite{DBLP:conf/kdd/Guo0CLSD23}. AAGOD employs a graph adaptive amplifier module, which is integrated into a well-trained GNN to facilitate graph OOD detection. Unlike the test-time training approach, AAGOD employs the training set for the OOD detection training process. As of the submission of this paper, AAGOD has not released the code, so we conduct comparisons based on the experimental results outlined in the AAGOD evaluations.
    AAGOD exists in two versions: AAGOD-GIN\ensuremath{_S+} and AAGOD-GIN\ensuremath{_L+}, which correspond to distinct OOD evaluation methods. Unreported experimental results are denoted by `$-$'. 
\end{itemize}

We employ a GIN \cite{DBLP:conf/iclr/XuHLJ19} as the well-trained GNN encoder. We use the Adam optimizer~\cite{kingma2014adam} for optimization.
All experimental procedures are conducted on a GeForce GTX TITAN X GPU device with $24$ GB memory. We repeated all experiments five times and report the average score and standard deviation. 
\begin{table*}[t]
\centering
\caption{Anomaly detection results in terms of AUC score ($\%$). The best results are highlighted with \textbf{bold}.} 
\label{ano}
\vspace{-0.3cm}
\resizebox{1\textwidth}{!}{
\begin{tabular}{l  cccc cc c c c}
\toprule
Method & PK-OCSVM & PK-iF & WL-OCSVM & WL-iF & InfoGraph-iF & GraphCL-iF  & GTrans & Ours & \textit{Improve}\\
\midrule
PROTEINS-full & $50.49{\scriptstyle\pm4.92}$ & $60.70{\scriptstyle\pm2.55}$ & $51.35{\scriptstyle\pm4.35}$ & $61.36{\scriptstyle\pm2.54}$ & $57.47{\scriptstyle\pm3.03}$ & $60.18{\scriptstyle\pm2.53}$ & $60.16{\scriptstyle\pm5.06}$ & $\mathbf{77.92{\scriptstyle\pm2.37}}$ & $28.37\%$ \\
ENZYMES       & $53.67{\scriptstyle\pm2.66}$ & $51.30{\scriptstyle\pm2.01}$ & $\mathbf{55.24{\scriptstyle\pm2.66}}$ & $51.60{\scriptstyle\pm3.81}$ & $53.80{\scriptstyle\pm4.50}$ & $53.60{\scriptstyle\pm4.88}$ & $38.02{\scriptstyle\pm6.24}$ & $52.33{\scriptstyle\pm4.74}$ & $-$\\
AIDS          & $50.79{\scriptstyle\pm4.30}$ & $51.84{\scriptstyle\pm2.87}$ & $50.12{\scriptstyle\pm3.43}$ & $61.13{\scriptstyle\pm0.71}$ & $70.19{\scriptstyle\pm5.03}$ & $79.72{\scriptstyle\pm3.98}$ & $84.57{\scriptstyle\pm1.91}$ & $\mathbf{95.50{\scriptstyle\pm0.99}}$ & $12.92\%$\\
DHFR          & $47.91{\scriptstyle\pm3.76}$ & $52.11{\scriptstyle\pm3.96}$ & $50.24{\scriptstyle\pm3.13}$ & $50.29{\scriptstyle\pm2.77}$ & $52.68{\scriptstyle\pm3.21}$ & $51.10{\scriptstyle\pm2.35}$ & $61.15{\scriptstyle\pm2.87}$ & $\mathbf{61.52{\scriptstyle\pm2.86}}$&  $0.60\%$\\
BZR           & $46.85{\scriptstyle\pm5.31}$ & $55.32{\scriptstyle\pm6.18}$ & $50.56{\scriptstyle\pm5.87}$ & $52.46{\scriptstyle\pm3.30}$ & $63.31{\scriptstyle\pm8.52}$ & $60.24{\scriptstyle\pm5.37}$ & $51.97{\scriptstyle\pm8.15}$ & $\mathbf{64.77{\scriptstyle\pm3.87}}$ & $2.31\%$\\
COX2          & $50.27{\scriptstyle\pm7.91}$ & $50.05{\scriptstyle\pm2.06}$ & $49.86{\scriptstyle\pm7.43}$ & $50.27{\scriptstyle\pm0.34}$ & $53.36{\scriptstyle\pm8.86}$ & $52.01{\scriptstyle\pm3.17}$ & $53.56{\scriptstyle\pm3.47}$ & $\mathbf{59.99{\scriptstyle\pm9.76}}$ &$12.01\%$\\
DD            & $48.30{\scriptstyle\pm3.98}$ & $71.32{\scriptstyle\pm2.41}$ & $47.99{\scriptstyle\pm4.09}$ & $70.31{\scriptstyle\pm1.09}$ & $55.80{\scriptstyle\pm1.77}$ & $59.32{\scriptstyle\pm3.92}$ & $76.73{\scriptstyle\pm2.83}$ & $\mathbf{77.62{\scriptstyle\pm2.88}}$ & $1.16\%$\\
NCI1          & $49.90{\scriptstyle\pm1.18}$ & $50.58{\scriptstyle\pm1.38}$ & $50.63{\scriptstyle\pm1.22}$ & $\mathbf{50.74{\scriptstyle\pm1.70}}$ & $50.10{\scriptstyle\pm0.87}$ & $49.88{\scriptstyle\pm0.53}$ & $41.42{\scriptstyle\pm2.16}$ & $45.96{\scriptstyle\pm2.42}$  &$-$\\
IMDB-B        & $50.75{\scriptstyle\pm3.10}$ & $50.80{\scriptstyle\pm3.17}$ & $54.08{\scriptstyle\pm5.19}$ & $50.20{\scriptstyle\pm0.40}$ & $56.50{\scriptstyle\pm3.58}$ & $56.50{\scriptstyle\pm4.90}$ & $45.34{\scriptstyle\pm3.75}$ & $\mathbf{65.46{\scriptstyle\pm4.34}}$ & $15.86\%$\\
REDDIT-B      & $45.68{\scriptstyle\pm2.24}$ & $46.72{\scriptstyle\pm3.42}$ & $49.31{\scriptstyle\pm2.33}$ & $48.26{\scriptstyle\pm0.32}$ & $68.50{\scriptstyle\pm5.56}$ & $71.80{\scriptstyle\pm4.38}$ & $69.71{\scriptstyle\pm2.21}$ & $\mathbf{80.31{\scriptstyle\pm0.85}}$ &$11.85\%$\\
COLLAB        & $49.59{\scriptstyle\pm2.24}$ & $50.49{\scriptstyle\pm1.72}$ & $\mathbf{52.60{\scriptstyle\pm2.56}}$ & $50.69{\scriptstyle\pm0.32}$ & $46.27{\scriptstyle\pm0.73}$ & $47.61{\scriptstyle\pm1.29}$ & $49.76{\scriptstyle\pm1.37}$ & $44.91{\scriptstyle\pm0.87}$ &$-$\\
HSE           & $57.02{\scriptstyle\pm8.42}$ & $56.87{\scriptstyle\pm10.51}$ & $62.72{\scriptstyle\pm10.13}$ & $53.02{\scriptstyle\pm5.12}$ & $53.56{\scriptstyle\pm3.98}$ & $51.18{\scriptstyle\pm2.71}$ & $58.49{\scriptstyle\pm2.68}$ & $\mathbf{63.05{\scriptstyle\pm0.90}}$ &$0.53\%$\\
MMP           & $46.65{\scriptstyle\pm6.31}$ & $50.06{\scriptstyle\pm3.73}$ & $55.24{\scriptstyle\pm3.26}$ & $52.68{\scriptstyle\pm3.34}$ & $54.59{\scriptstyle\pm2.01}$ & $54.54{\scriptstyle\pm1.86}$ & $48.19{\scriptstyle\pm3.74}$ & $\mathbf{69.41{\scriptstyle\pm0.04}}$ & $25.65\%$\\
p53           & $46.74{\scriptstyle\pm4.88}$ & $50.69{\scriptstyle\pm2.02}$ & $54.59{\scriptstyle\pm4.46}$ & $50.85{\scriptstyle\pm2.16}$ & $52.66{\scriptstyle\pm1.95}$ & $53.29{\scriptstyle\pm2.32}$ & $53.74{\scriptstyle\pm2.98}$ & $\mathbf{63.27{\scriptstyle\pm0.04}}$ & $15.90\%$\\
PPAR-gamma    & $53.94{\scriptstyle\pm6.94}$ & $45.51{\scriptstyle\pm2.58}$ & $57.91{\scriptstyle\pm6.13}$ & $49.60{\scriptstyle\pm0.22}$ & $51.40{\scriptstyle\pm2.53}$ & $50.30{\scriptstyle\pm1.56}$ & $56.20{\scriptstyle\pm1.57}$ & $\mathbf{68.23{\scriptstyle\pm1.54}}$ & $17.82\%$\\
\midrule
\textit{Avg. Rank}    & $6.3$ & $5.9$ & $4.4$ & $5$ & $4.2$ & $4.3$ & 4.4 & $\mathbf{2.1}$  & \\
\bottomrule
\end{tabular}}
\end{table*}
\begin{table*}[t]
\centering
\caption{Ablation study results in terms of AUC score ($\%$). The best results are highlighted with \textbf{bold}.} 
\label{abl}
\vspace{-0.3cm}
\resizebox{1\textwidth}{!}{
\begin{tabular}{ccc cccccccccc}
\toprule
\multirow{2}{*}{\scalebox{1.3}{$\mathcal{L}_{s}$}} & \multirow{2}{*}{\scalebox{1.3}{$\mathcal{L}_{m}$}} & \multirow{2}{*}{\scalebox{1.3}{$\mathcal{L}_{d}$}} & BZR & PTC-MR & AIDS & ENZYMES & IMDB-M & Tox21 & FreeSolv & BBBP & ClinTox & Esol \\

\cmidrule{4-13}
  &   &   & COX2 & MUTAG & DHFR & PROTEIN & IMDB-B & SIDER & ToxCast & BACE & LIPO & MUV \\
\midrule
\checkmark & -& - & $39.10{\scriptstyle\pm1.59}$ & $67.49{\scriptstyle\pm3.48}$ & $72.28{\scriptstyle\pm1.03}$ & $61.55{\scriptstyle\pm3.25}$ & $79.01{\scriptstyle\pm0.02}$ & $68.89{\scriptstyle\pm0.01}$ & $68.58{\scriptstyle\pm0.10}$ & $65.20{\scriptstyle\pm1.06}$ & $56.50{\scriptstyle\pm0.16}$ & $55.70{\scriptstyle\pm1.38}$  \\
- & \checkmark& - & $73.53{\scriptstyle\pm0.04}$ & $72.70{\scriptstyle\pm0.04}$ & $82.34{\scriptstyle\pm0.18}$ & $52.74{\scriptstyle\pm0.01}$ & $69.98{\scriptstyle\pm0.02}$ & $68.87{\scriptstyle\pm0.01}$ & $67.65{\scriptstyle\pm0.08}$ & $63.77{\scriptstyle\pm0.24}$ & $66.71{\scriptstyle\pm0.05}$ & $85.82{\scriptstyle\pm0.08}$  \\
- & -& \checkmark & $72.11{\scriptstyle\pm0.28}$ & $72.72{\scriptstyle\pm0.06}$ & $92.52{\scriptstyle\pm0.02}$ & $51.95{\scriptstyle\pm0.11}$ & $79.02{\scriptstyle\pm0.02}$ & $68.90{\scriptstyle\pm0.01}$ & $68.62{\scriptstyle\pm0.06}$ & $76.98{\scriptstyle\pm0.04}$ & $54.03{\scriptstyle\pm0.04}$ & $80.23{\scriptstyle\pm0.05}$  \\
\checkmark & \checkmark& - & $\mathbf{83.06{\scriptstyle\pm1.45}}$ & $81.83{\scriptstyle\pm0.12}$ & $96.33{\scriptstyle\pm0.29}$ & $65.67{\scriptstyle\pm2.31}$ & $79.02{\scriptstyle\pm0.03}$ & $68.92{\scriptstyle\pm0.01}$ & $68.77{\scriptstyle\pm0.10}$ & $76.65{\scriptstyle\pm0.25}$ & $63.25{\scriptstyle\pm0.56}$ & $85.90{\scriptstyle\pm0.44}$  \\
\checkmark & -& \checkmark & $48.22{\scriptstyle\pm3.09}$ & $67.09{\scriptstyle\pm2.58}$ & $70.64{\scriptstyle\pm3.28}$ & $60.46{\scriptstyle\pm3.40}$ & $79.02{\scriptstyle\pm0.02}$ & $68.89{\scriptstyle\pm0.01}$ & $68.67{\scriptstyle\pm0.14}$ & $66.21{\scriptstyle\pm0.89}$ & $56.34{\scriptstyle\pm0.17}$ & $54.60{\scriptstyle\pm1.29}$  \\
- & \checkmark& \checkmark & $73.58{\scriptstyle\pm0.04}$ & $72.73{\scriptstyle\pm0.05}$ & $82.83{\scriptstyle\pm0.20}$ & $52.73{\scriptstyle\pm0.02}$ & $79.01{\scriptstyle\pm0.02}$ & $68.88{\scriptstyle\pm0.01}$ & $67.73{\scriptstyle\pm0.07}$ & $63.70{\scriptstyle\pm0.14}$ & $\mathbf{66.76{\scriptstyle\pm0.08}}$ & $85.76{\scriptstyle\pm0.09}$  \\
\midrule
\checkmark & \checkmark& \checkmark & $82.16{\scriptstyle\pm0.15}$ & $\mathbf{81.84{\scriptstyle\pm0.57}}$ & $\mathbf{96.43{\scriptstyle\pm0.25}}$ & $\mathbf{66.29{\scriptstyle\pm1.54}}$ & $\mathbf{79.03{\scriptstyle\pm0.03}}$ & $\mathbf{68.92{\scriptstyle\pm0.01}}$ & $\mathbf{68.83{\scriptstyle\pm0.02}}$ & $\mathbf{77.07{\scriptstyle\pm0.03}}$ & $62.46{\scriptstyle\pm0.54}$ & $\mathbf{85.91{\scriptstyle\pm0.27}}$  \\
\bottomrule
\end{tabular}}
\vspace{-0.1cm}
\end{table*}

\subsection{Performance of Graph OOD Detection}
We conduct a comparative analysis of our proposed method against $13$ competing approaches across $10$ graph OOD detection datasets. The results, in terms of AUC scores, are summarized and presented in Table \ref{oodd}. From this comprehensive comparison, we garner the following insights: 1) GOODAT showcases a remarkable performance by outperforming all baseline methods on $8$ datasets. Furthermore, when considering the average rank across all methods, our proposed approach stands as the leader. This underscores the effectiveness of GOODAT in accurately detecting OOD samples within diverse graph-structured datasets. 2) While we may not have achieved the absolute best results on two datasets, our performance is in close proximity to the optimum results. One plausible explanation for this outcome is the relatively high edge density observed in these datasets, which potentially reduces the effect of GIB-boosted loss for subgraph compression. 3) Although the test-time graph OOD generalization method GTrans is also effective for OOD detection, our method has achieved overall advantages on all datasets. This demonstrates that directly substituting of graph OOD generalization for graph OOD detection may lead to sub-optimal performance. 4) Compared to the data-centric method, AAGOD, our proposed approach shows superiority across all datasets. This implies that our method achieves better results only using a test dataset, showcasing enhanced efficacy.

\subsection{Performance of Graph Anomaly Detection}

To assess the potential applicability of GOODAT on graph anomaly detection tasks, we conduct experiments on $15$ datasets, following the evaluation protocol in \cite{DBLP:conf/wsdm/LiuD0P23, DBLP:conf/wsdm/MaPCH22}. We select $6$ graph anomaly detection methods and a test-time training method as baselines, and the experiment results are summarized in Table \ref{ano}. The AAGOD baseline is not included here as its code is not released. Experimental results indicate that GOODAT's applicability extends seamlessly to the anomaly detection scenario, showcasing remarkable performance. This superior performance can be attributed to the inherent strengths of our method, which effectively enhances the distinctions in anomalous samples by utilizing GIB-boosted losses with ID-label guidance. Moreover, we observe that GOODAT exhibits significant advantages in anomaly detection over the use of GTrans. This observation emphasizes the universality of GOODAT in contrast to other test-time-oriented techniques.

\subsection{Ablation Study}
GOODAT incorporates three core loss functions: subgraph GIB loss $\mathcal{L}_{s}$, masked graph GIB loss $\mathcal{L}_{m}$, and graph distribution separation loss $\mathcal{L}_{d}$. To evaluate the effectiveness of each of them, an ablation study is conducted and the results are summarized in Table \ref{abl}, where a `$\checkmark$' indicates the presence and a `$-$' denotes the absence of a component. As a result of analyzing the table, several key observations emerge. Firstly, utilizing all components concurrently yields optimal results on $8$ out of $10$ datasets, with decent results on the remaining $2$ datasets. This demonstrates the effectiveness of integrating multiple loss functions to enhance graph OOD detection. Secondly, the distinct contributions of each loss function underscore their effectiveness as individual components. Finally, combining the two loss functions often leads to performance improvements, outperforming isolated evaluations of individual components.

\subsection{Parameter Sensitivity Analysis}
In GOODAT, two hyperparameters $\alpha$ (in Eq. \ref{ls}) and $\beta$ (in Eq. \ref{lm}) are employed to control the involvement degree of subgraph GIB loss and masked graph GIB loss, respectively. We conduct a parameter sensitivity experiment on the PTC-MR/MUTAG dataset, where $\alpha$ is selected from $\{0.1, 0.3, 0.5, 0.7, 0.9\}$ and $\beta$ is selected from $\{0.01, 0.03, 0.05, 0.07, 0.09\}$. The results are depicted in Fig. \ref{vis} (a). 
The results reveal that, when $\beta$ is fixed, optimal outcomes are achieved with $\alpha$ values within the range of $0.1$-$0.3$. This implies that a slight level of compression on subgraphs proves the most effective results. Likewise, when $\alpha$ is held constant, $\beta$ values in the range of $0.3$-$0.5$ yield optimal results. This observation suggests that moderate compression of masked graph components enhances model effectiveness.

\subsection{Visualization}
\begin{figure}
	\centering
	\small
        \subfigure[{Influence on the parameter $\alpha$ and $\beta$.}]{\includegraphics[width=0.35\columnwidth]{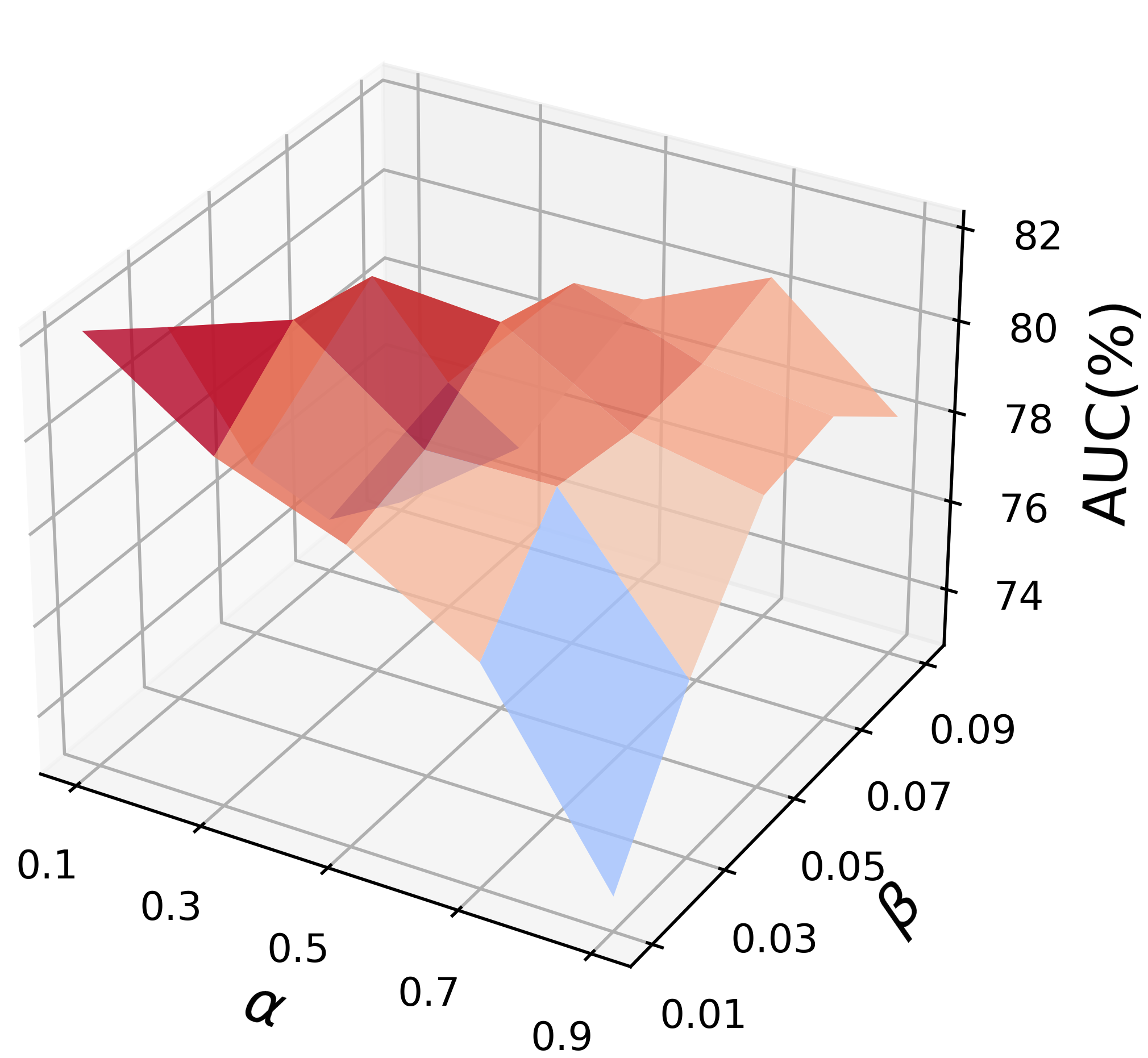}}
	\subfigure[{OOD and ID G-Emb.}]{\includegraphics[width=0.3\columnwidth]{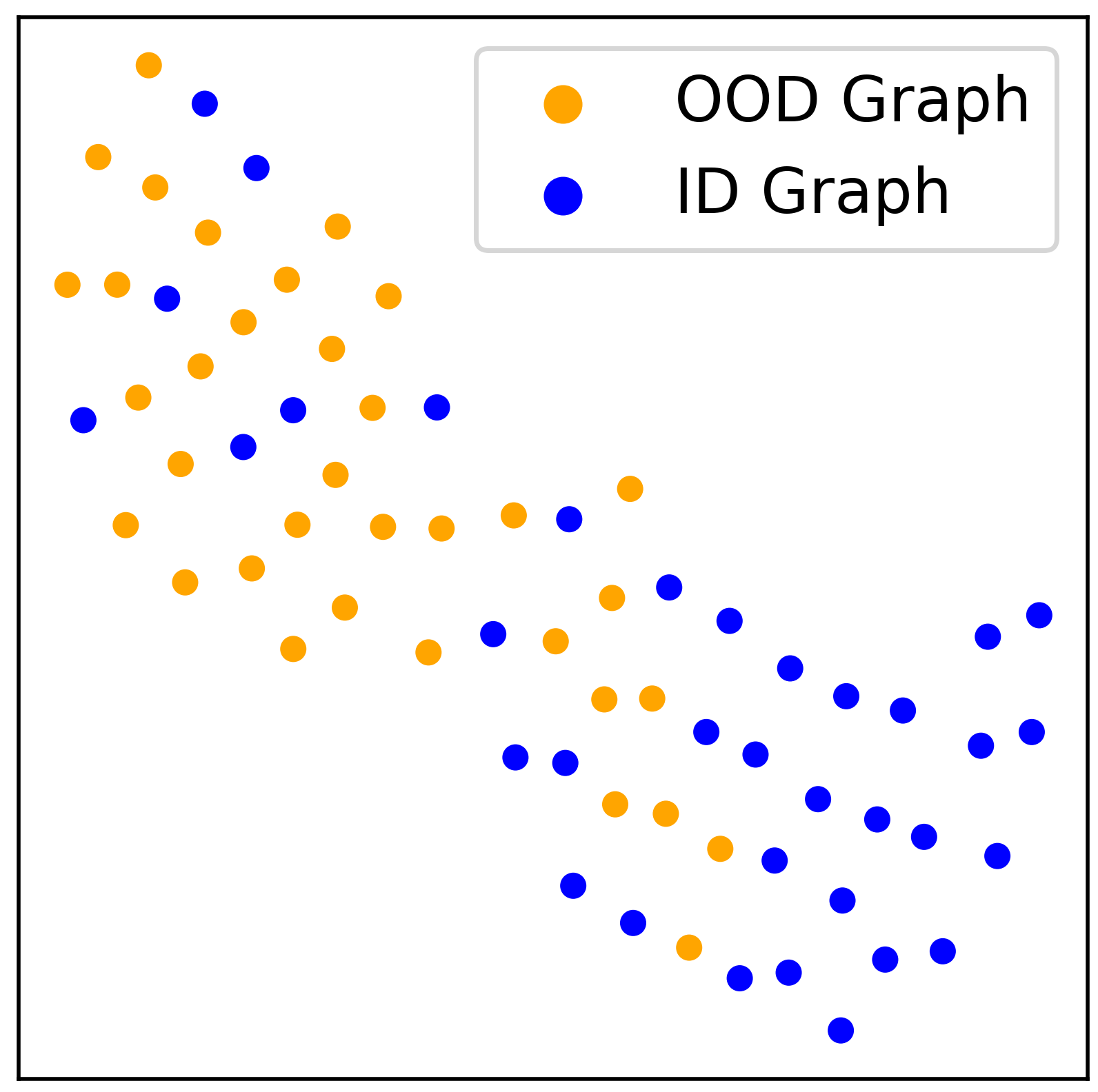}}
	\subfigure[{Sub and Masked G-Emb.}]{\includegraphics[width=0.3\columnwidth]{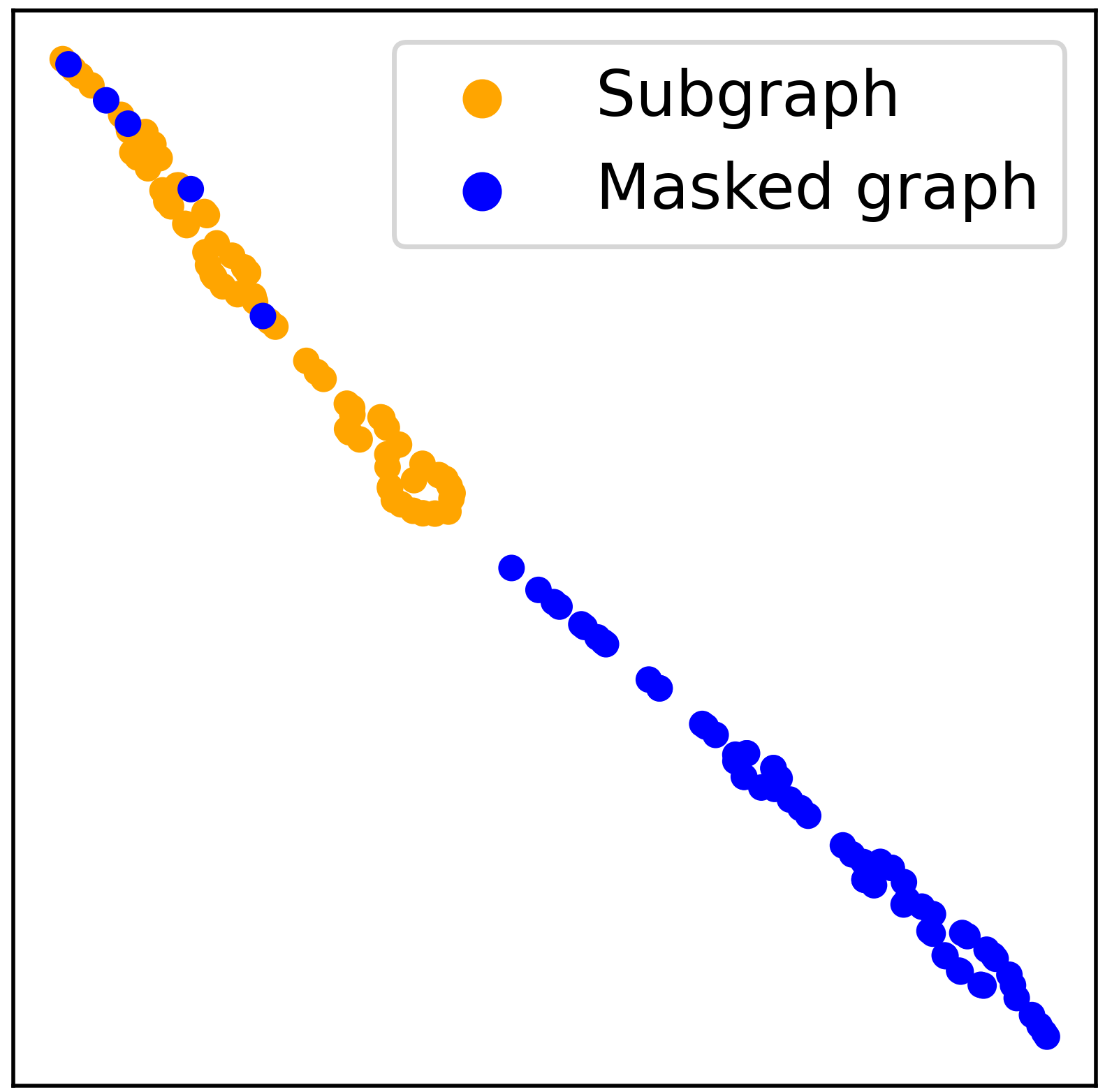}}
	\caption{Parameter sensitivity analysis and visualization.}
	\label{vis}
\end{figure}

To visually demonstrate the impact of our methods, we visualize distributions of subgraph embeddings to show the distinction between ID and OOD graphs. Fig. \ref{vis} (b) shows the embedding distributions of OOD subgraphs and ID subgraphs. We observe that the ID subgraph and OOD subgraph can be clearly distinguished, which indicates that GOODAT can detect OOD graphs intuitively. Fig. \ref{vis} (c) shows the distribution of subgraphs (i.e., $Z$) and masked graphs (i.e., $Z'$), where the distinction between subgraphs and masked graphs is also obvious. This demonstrates the effectiveness of our proposed graph distribution separation loss.

\section{Conclusions}
In this paper, we make the first attempt toward detecting graph out-of-distribution (OOD) samples at test time. To achieve this, we introduce a pioneering method, named GOODAT, which is a data-centric, unsupervised, and plug-and-play solution. With a graph masker applied to the input test graph, GOODAT identifies the clear differentiation between OOD graphs and ID graphs. We design three GIB-boosted losses to optimize the graph masker. Comprehensive experimentation demonstrates the superiority of GOODAT compared to baseline methods across diverse real-world benchmark datasets.

\section{Acknowledgments}
This work is supported by National Key Research and Development Program of China (2023YFC3304503), National Natural Science Foundation of China (92370111, 62276187, 62272340).

\bigskip
\bibliography{aaai24}

\end{document}